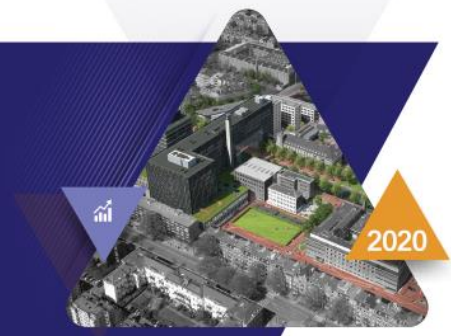

# A Review into Data Science and Its Approaches in Mechanical Engineering

**Ashkan Yousefi Zadeh \*, Meysam Shahbazy**

1.  Msc stundent, Mechanical engineering department, University of Guilan,
    a-yousefizadeh@msc.guilan.ac.ir
2.  Msc stundent, Mechanical engineering department, University of Guilan,
    meysamshahbazy@msc.guilan.ac.ir

## Abstract

Nowadays it is inevitable to use intelligent systems to improve the performance and optimization of different components of devices or factories. Furthermore, it's so essential to have appropriate predictions to make better decisions in businesses, medical studies, and engineering studies, etc. One of the newest and most widely used of these methods is a field called 'Data Science' that all of the scientists, engineers, and factories need to learn and use them in their careers. This article briefly introduced data science and reviewed its methods, especially it's usages in mechanical engineering and challenges and ways of developing data science in mechanical engineering. In the introduction, different definitions of data science and its background in technology reviewed. In the following, data science methodology which is the process that a data scientist needs to do in its works been discussed. Further, some researches in mechanical engineering area that used data science methods in their studies, are reviewed. Eventually, it has been discussed according to the subjects that have been reviewed in the article, why it is necessary to use data science in mechanical engineering researches and projects.

## 1. Introduction

In 1960 a Danish computer scientist called Peter Naur used the word "datalogy", which, in his country, was considered an alternative name for computer science[1]. He defined that using and understanding the foundations of computer science as "the science of the nature and use of data", is called 'datalogy'[1]. In 1962 John Tukey (American mathematician) published an article named "The Future of Data Analysis", in this article he has been hinted into mathematical-statistics evolve, He declared that the research process was focused and wrong at the time and could be ineffective and even destructive so the research field of statistics needed to be developed and redirected[2]. In 1966 Peter Naur published a booklet named "Plan for a course in datalogy and datamatics", He explained fundamentals of datalogy, theories of data process, comprising data, data representations, and he has been giving a clear consideration of these principles[1]. Gil Press wrote the article "A Very Short History Of Data Science" at Forbes, he mentioned: In 1977, Tukey published Exploratory Data Analysis, arguing that more emphasis needed to be placed on using data to suggest hypotheses to test and that Exploratory Data Analysis and Confirmatory Data Analysis "can—and should proceed side by side."[3]. So these two are considered as the founders of data science, during the years many ideas and thoughts about "Data Science" have been provided by many scientists[3]. There are many definitions for "Data Science" that several notable of them have been reviewed in the following. Waller and Fawcett believe: "Generally, data science is the application of quantitative and qualitative methods to solve relevant problems and predict outcomes."[4]. Murtaza Haider declared in his book: "I define data science as something that data scientists do.", also, he has been defined a data scientist as someone: "who finds solutions to problems by analyzing big or small data using appropriate tools and then tells stories to communicate her findings to the relevant stakeholders."[5]. The Data Science Association has provided some definitions: "Data" means a tangible or electronic record of raw (factual or non-factual) information (as measurements, statistics or information in numerical form that can be digitally transmitted or processed) used as a basis for reasoning, discussion, or calculation and must be processed or analyzed to be meaningful. It's also written "Data Science" means the scientific study of the creation, validation, and transformation of data to create meaning and "Data Scientist" means a professional who uses scientific methods to liberate and create meaning from raw data[6]. Tierney and Kelleher have described: "Data science encompasses a set of principles, problem definitions, algorithms, and processes for extracting non-obvious and useful patterns from large data sets."[7]. All of these definitions overlap, so to summarize it can be said that data science is to penetrate the depth of data and find its various elements and use them to make things easier to understand so the person doing this job is called a "data scientist". Now, according to the topics mentioned, the question may arise as to what is the difference between data science and statistics?! Or even what is the role of computer science in the meantime?! In "Figure 1", the overlap of mathematics and statistics with data science is well illustrated, and the position and role of computer science and machine learning are evident. It is clear that these issues are highly intertwined and it is impossible to distinguish them from one another. In this regard, Kharkovyna has used an interesting interpretation in his article: "A Data Scientist is one who knows more statistics than a programmer and more programming than a statistician."[8].

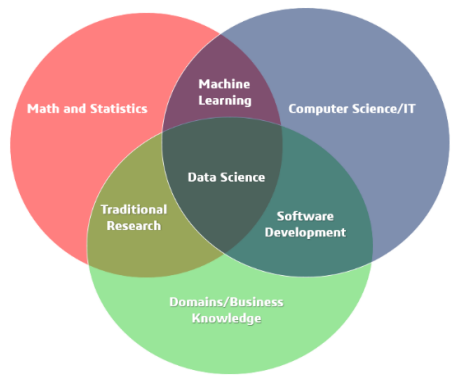

**Figure 1. Overlap of mathematics and statistics With data science [8].**

Based on "trends.google.com" as it's shown in "figure 2", there is a comparison between the rate of web searches about, "Data Science"," Statistic" and "Big Data" in the world during 2005-2020. As it is clear, the search for Statistics has been almost constant, in other hands Data Science and Big Data has been uptrend especially from 2011.Based on Google hints :" Numbers represent search interest relative to the highest point on the chart for the given region and time. A value of 100 is the peak popularity for the term. A value of 50 means that the term is half as popular. A score of 0 means there was not enough data for this term."

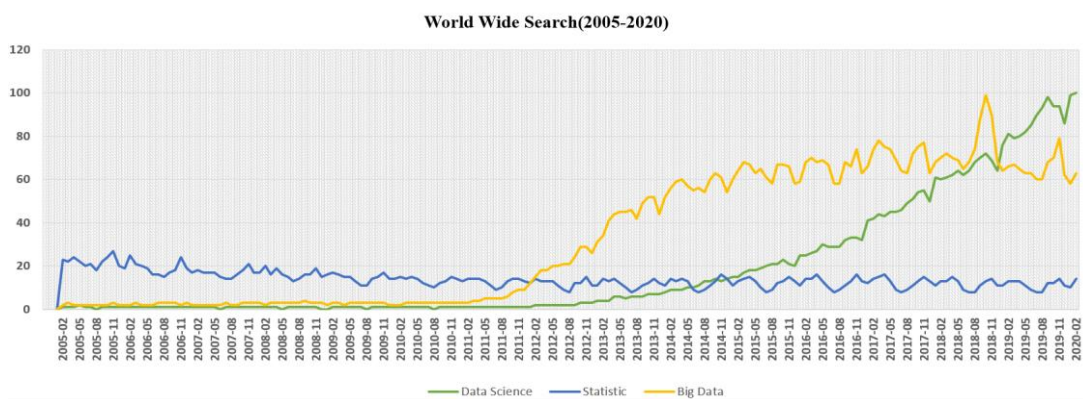

**Figure 2. Online search interest trends on data science related keywords by Google.**
**Note: The data was collected on February 2020.**

"Figure3" shows the results of a survey from 2,233 people that cognitiveclass.ai website published it in 2017. The question asked by the participants was:" What's your level of interest for the following technologies?"

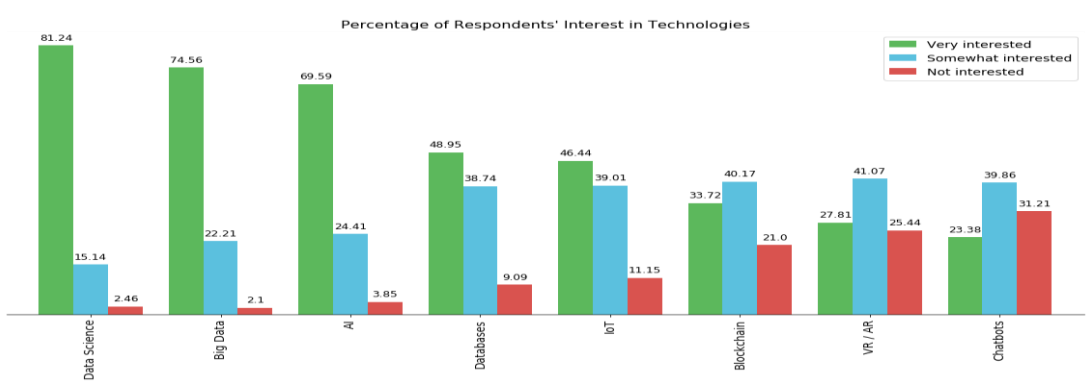

**Figure 3.[9]**

As it's obvious, the rate of being interested in Data Science is rather than others.

## 2 .The relationship between Data Science and Mechanical Engineering

In 2012, Davenport and Patil published an article that became very popular in a short time[10].when a mechanical engineer reads that article maybe ask itself, Can I become a "data scientist"?, answering this question is difficult, because they're both have their particular role in science and technology and comparing them together is inappropriate. Many foundations of mechanical engineering have been valid for many years, and even their history may be back to a hundred years ago, subjects like Kalman Filter, PD control, PID control, Tresca criterion, Duffing equation, Second law of thermodynamics, etc. [11]. Nowadays, the main problems of mechanical engineers are that they usually limited set of tools: there are benchmarks of standard tools for machine control, hardware design, etc. [11]. Of course, today there is some very good software such as MATLAB, CATIA, Abaqus, ANSYS, etc., that are very helpful to mechanical engineers, But imagine in a project something gone wrong, and the engineers and managers didn't identify that yet, everyone knows that the technologies used in mechanical engineering are largely complex, and it is not easy to find the relationships between the various factors and debugging them and solve the problems quickly is almost impossible, Then it becomes a disaster, and the result is a huge loss of financial, time, energy, manpower and so on. So maybe if there were a way to make online updating for appropriate predictions, analysis, and visualizations about the performance of equipment, parts, environmental conditions, and thousands of other factors that could affect the workflow, these circumstances never happen again. Given the explanations given, it can be said that the best solution is to use data science methods. As mentioned, data science depends on computer science, mathematics, and statistics. So those who have become mechanical engineers, they are definitely always familiar with these topics and can easily learn advanced topics that may be needed to learn data science, therefore mechanical engineers can operate completely across these fields[12]. In the last decades, the use of optimization algorithms, especially evolutionary algorithms, which one of the most well-known of them is the genetic algorithm, has increased among mechanical engineers. One of the advantages of data science methods is that they can be combined with optimization algorithms that can produce interesting results. So using data science will be a delight for scientists and engineers who have had experience in optimization. The following briefly has explained why Evolutionary Algorithms can be used in data science. There are many types of these algorithms that really are helpful, Evolutionary Algorithms are optimization algorithms based on haphazard population which always needs this three main elements as shown in "Figure 4", [13]. The following features make it possible to use evolutionary algorithms in data science:

- In the high dimensional search spaces, Evolutionary Algorithms work properly, while other algorithms are only doing optimizing.
- Evolutionary Algorithms can optimize a number of criteria at once.

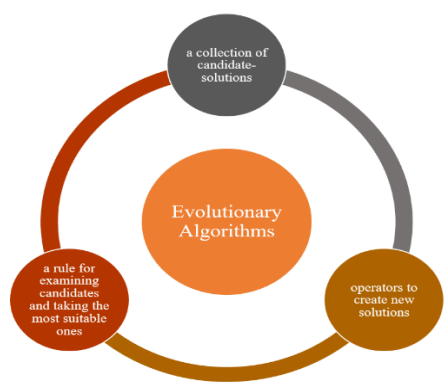

**Figure 4. Evolutionary Algorithms main elements**

This two features makes it possible to use "Feature Selection" to encounter data sets with thousands of attributes [13].

- Instance selection has been developed for population-based methods.
- Instance selection might be easily inserted in training a model.

And this two features makes it possible to Instance selection [13].

As a result, according to the material said, scientific data science is very extensive, and certainly, given the complex topics in physics and mechanics, using the benefits of data science can be very useful.

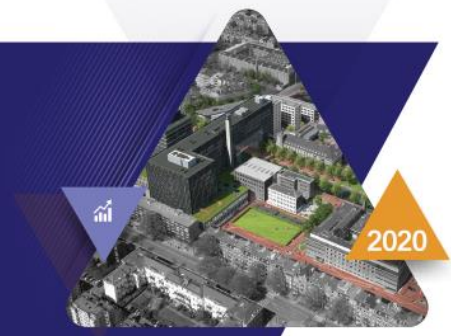

In the following, it's been discussed the main foundations of data science and after that has been provided examples of the use of data science techniques in some fields of mechanical engineering.

Today there are various companies and websites that hold competitions in the field of data science. Different companies and factories on these web sites express their needs in the field of data science in the form of competitions and data scientists from all around the world are enrolling in these competitions to solve these companies' problems. "kaggle.com" is one of the most popular of these websites. For example, in 2017 the Mercedes-Benz company held a contest on the site with this title: "Can you cut the time a Mercedes-Benz spends on the test bench?", and they requested the participants to tackle the curse of dimensionality and reduce the time that cars spend on the test bench. They mentioned that winning algorithms will contribute to speedier testing, resulting in lower carbon dioxide emissions without reducing Daimler's standards[14].

Also, in 2017 BOSCH Company held a contest on the site with this title:" Reduce manufacturing failures", and they requested the participants to predict internal failures using thousands of measurements and tests made for each component along the assembly line. They mentioned, this would enable Bosch to bring quality products at lower costs to the end-user[15].

## 3 .Data Science Foundations

In this section, the various components of the process used in data science are briefly described, "Figure 5" illustrates the general trend of these processes.

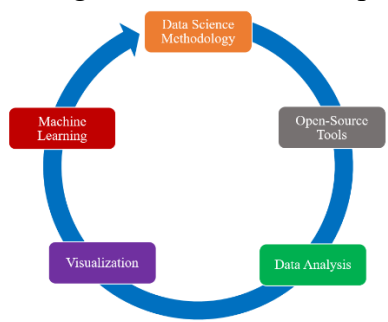

**Figure 5. The general trend of Data Science**

### 1-3- Data Science Methodology:

The first and most important step of Data science methodology is to have an understanding of the issue and Clarification and having a good vision about the situation. This comprehension lets the data scientist, know what stage it is, and what stage it wants to reach [16]."Figure 6" shown the methodology process:

**Figure 6. Data Science Methodology Process [17]**

John Rollins (Data Scientist), believes: "having a clearly defined question is vital because it ultimately directs the analytic approach that will be needed to address the question."[16].

- Business understanding: All projects and researches notwithstanding of their size, begin with business understanding which this step is the toughest one[17]. The question that needs to be answered at this point is what is the problem the data scientist wants to solve?[16].

- Analytic approach: after the problem clarification has done, the data scientist must start using analytic methods and solve the problem in this way[17]. Actually, in this section, the main work is to identify required and appropriate patterns, which their results in the problem will be the best[16]. Reach these purposes is by using statistical and machine learning techniques[17].

- Data requirements: Obviously, to have a proper analytical approach, it needs to have reliable and useful data.

- Data collection: In this section, collecting the data and select the appropriate data that is needed e.g.: structured, unstructured and semi-structured that are related to the problem domain must be done[17].

- Data understanding: this section includes all actions correlated to creating the data set and the question it's must be answered is, the data that are collected can solve the problem?![16].

Data preparation: this step must include actions like data cleaning, combining data from multiple sources and transforming data into more useful variables that are essential to building the data set which will be used in the modeling step. There is sometimes possible, data scientist needs to do feature engineering and text analytics to determine new structured variables, for the set of predictors and also improving the model's accuracy [17], it is necessary to take appropriate time in this section and use the tools available to automate common Steps to accelerate data preparation [16].

- Modeling: as it's shown in the figure, this step may have most iterate, Modeling include making training set from the initial dataset or historical data to generate predictive or descriptive models using the analytic method[17].

- Evaluation: in this step performance of the model must be evaluated by using some tools and methods such as tables and graphs or using a testing set for a predictive model[17].

- Deployment: After making an evaluated model that can be approved, it is time to run the model in a production environment or a comparable test environment[17].

- Feedback: after obtaining results from the executed model, the final step is to getting feedback on the model's performance and observing how it affects its deployment environment. This step allows the data scientist to refine and improve the model if it's needed[17].

The end of this section hint at the explicit and implicit analytical spectrum of evolution and the relationship between explicit and implicit analysis, that being familiar with these concepts helps to handle the first task of the methodology that is business understanding, "Figure 7" shows a high-level conceptual view of the spectrum and evolution of analytical components and tasks in terms of two major dimensions [18].

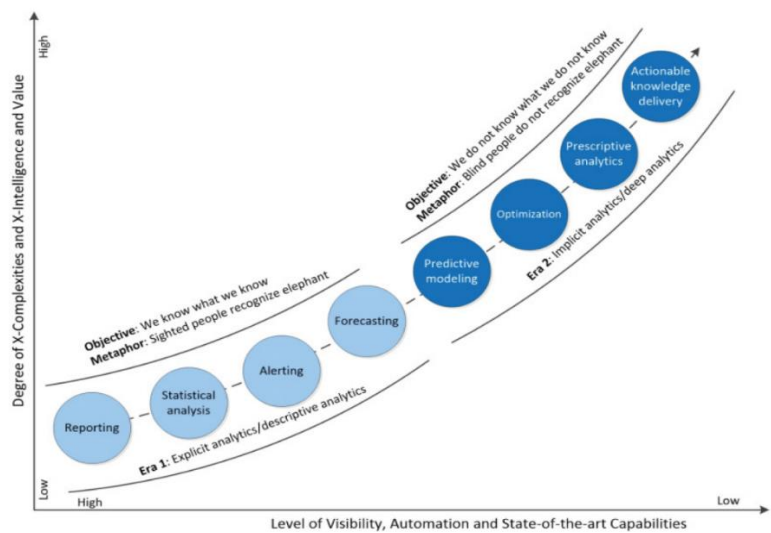

**Figure 7. Explicit to implicit analytics spectrum and evolution [18].**

### 2-3- Data Analysis:

One most highly required skills that a data scientist needs to have, is domination one the deep domain knowledge and an extensive set of analytical skills.

- Descriptive Data Analytics: This is an initial step for data processing, it can be the creator of the context of historical data to obtain useful data, and therefore it allows managing the data for advanced analysis. Descriptive Analytics starts by providing a static view of the past[20]. In other words Descriptive analytics represents the use of a range of historic data to draw collation[21]. Descriptive Analytics trying to explain and exhibit data in a form that can be simply comprehended by a wide type of business readers[22]. In short form, the concept in this method is to find out, what is happening?[19].

- Diagnostic Analytics: This method allows data scientists to understanding data more quickly therefore that causes answer critical questions faster[23]. The function and application of this type of analysis are generally divided into three categories: 1- Identifying anomalies, 2- Penetrating to the analytics (discovery), 3- Discovering causal connections[24]. In short form, the concept in this method is to find out, why is it happening?[19].

- Predictive Analytics: This method with applying historical and current data on the various type of predicting techniques, shows the prediction of future results. It should also be noted that predictive analytics methods are fundamentally based on statistical methods[25]. In short form, the concept in this method is to find out, what is likely to happen?[19].

- Prescriptive Analytics: This methods are recognized as the next boundary in the field of business analytics. These analytics methods are considered as the new business analytics sample[26]. Also, these analytics methods can be considered as a type of predictive analytics and it works towards the best appropriate term of actions among given parameters. Actually it's not a specific analytical technique but with using and combining the previous three methods helps to data scientists and businesses to make the best and proper decisions[20]. In short form, the concept in this method is to find out, what do I need to do?[19].

As mentioned at the beginning of this section, these analytics have many methods and they can implement and develop in many environments. In the meantime, Python and R, are programming languages that have very suitable and useful libraries and packages for these tasks.

## 4 . Applications of Data Science to Mechanical Engineering

In the following, some examples of applications and uses of data science in mechanical engineering which scientists and engineers have done in their researches and works are briefly discussed and results of them have been provided:

### 1-4- An Example of Biomechanical applications

Nasrabadi and colleagues published an article in 2019, named "Optimal Sensor Configuration for Activity Recognition during Whole-body Exercises". This research was to design an affordable and accurate wearable device with IMUs (Inertial Measurement Units) to detect thirty-four different motor activities in a customized training program called LSVT-BIG1 (Lee Silverman Voice Technique-Big), which is usually used for people with PD. They used Dimension Reduction in their research by using the PCA algorithm, NM (Nearest Mean), RBF (Radial Basis Function), SVM, and K-NN classifiers to trained and used to recognize the activity[27].

"Table 1" shows their results of the first evaluation with different classifiers.

**Table 1.[27]**

| Classifiers | NM | k-NN | RBF | SVM |
|---|---|---|---|---|
| Accuracy | 1 | 1 | 0.99 | 0.97 |
| Recall | 1 | 1 | 0.97 | 0.93 |
| Precision | 1 | 1 | 0.98 | 0.91 |
| Specificity | 1 | 1 | 0.99 | 0.96 |
| F1-Score | 1 | 1 | 0.97 | 0.92 |
| Computation Duration, days | 1.5 | 6 | 18 | 33 |

Their results show despite the expected performance of RBF and SVM classifiers, they were not able to identify activities efficiently in that study. They found out optimizing the parameters of SVM (i.e. criteria limit and sigma) for each stage of classification can lead to better performance. They also concluded that adjusting the number of periods for each RBF trained for classification may result in a more accurate and quicker recognition[27].

### 2-4- An Example of Robotic applications

Pane and colleagues published an article in 2019, named "Reinforcement learning based compensation methods for robot manipulators".

They developed and implemented two Reinforcement learning based compensation schemes to improve the suboptimal tracking performance of a feedback controller in a multi DOF robot arm[28]. They introduced two reinforcement learning (RL) based compensation methods. The learned correction signal, which compensated for unmolded aberrations, was added to the existed nominal input to enhance the control performance. The proposed learning algorithms were been evaluated on a 6-DoF industrial robotic manipulator arm to follow different kinds of reference paths, such as a square or a circular path, or to track a trajectory on a three-dimensional surface. The first method was compensating the control input given by the nominal controller whereas the second method was compensating the nominal reference trajectory.

They found out that the Reinforcement learning control input compensation method has an advantage in a faster response since it compensates in the velocity space, thus a higher bandwidth is obtained and it also achieves a smaller error compared to the second method.

"Table 2" shows their results for the $z$-axis tracking performance of RL compensation methods compared to the other controllers for square reference tracking. "Table 3" shows their results for the Tracking performance comparison for the circular reference.

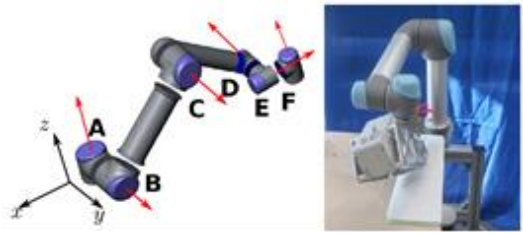

**Figure 8. The UR5 Robot. Left: The joint's axis positioning and the reference frame used in their research paper. The joints in alphabetical order (A to F): base, elbow, shoulder, wrist 1, wrist 2, and wrist 3. Picture is courtesy of Universal Robots. Right: the 3D printing system with the robot moving on top of the surface of a curved object [28].**

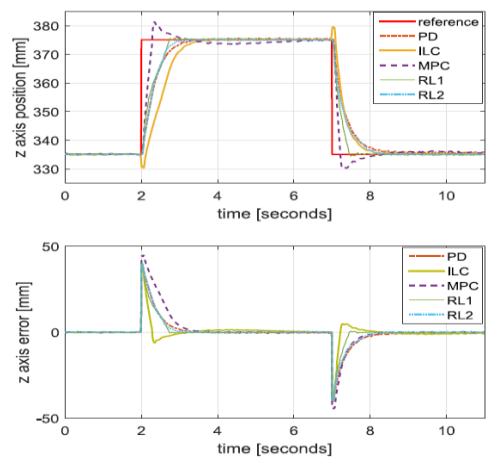

**Figure 9. The tracking performance of the proposed RL compensation methods in their research compared to the benchmark controllers for the square reference [28].**

**Table 2.[28]**

| Error (mm) | PD | MPC | ILC | RL-1 | RL-2 |
|---|---|---|---|---|---|
| Final steady state | −0.5858 | **0.0185** | −0.4798 | 0.0412 | 0.1423 |
| RMS | 7.4669 | 9.676 | **5.502** | 6.4721 | 7.2051 |

**Table 3.[28]**

| Error measure (mm) | MPC | ILC | PD | RL-1 | RL-2 |
|---|---|---|---|---|---|
| RMS $x$ | 1.0613 | 0.5109 | 4.2388 | **0.4847** | 0.4962 |
| RMS $y$ | 1.0108 | 0.4662 | 1.9859 | 0.3215 | **0.2408** |
| Max absolute $x$ | 1.6935 | 1.0565 | 6.0253 | 1.0529 | **0.9557** |
| Max absolute $y$ | 1.6938 | 1.9269 | 3.0798 | **1.1213** | 1.6508 |

"Table 4" shows their results for the tracking performance comparison for the 3D printing reference.

**Table 4.[28]**

| Error (mm) | MPC | ILC | PD | RL-1 | RL-2 |
|---|---|---|---|---|---|
| RMS $x$ | 1.9287 | 0.6111 | 19.3509 | **0.46153** | 0.82856 |
| RMS $y$ | **0.0616** | 0.0979 | 0.53016 | 0.20632 | 0.1972 |
| RMS $z$ | **0.3107** | 0.3440 | 2.6643 | 0.36981 | 0.43733 |
| Max absolute $x$ | 2.3499 | 3.9082 | 20.8915 | **1.2626** | 1.9246 |
| Max absolute $y$ | **0.2963** | 0.4854 | 2.0685 | 0.60674 | 0.5844 |
| Max absolute $z$ | 1.1019 | 1.5896 | 4.2995 | 1.3896 | **1.0341** |

### 3-4- An Example of Control engineering applications

Chaibakhsh and colleagues published an article in 2016, named "Early fault detection in transient conditions for a steam power plant subsystem using support vector machine". They used a support vector machine (SVM) method for early fault detection in a Benson type once-through boiler. Appropriate fault model was developed using SVM with radial basis function (RBF) as the kernel. The performance of the fault detection system was evaluated with respect to the similar faults at two different time periods that happen in a steam power plant. The troubleshooting system they introduced using this method worked well and it is capable of reducing the probability of boiler exiting from its workspace,     reducing maintenance costs, Increasing accessibility to the power plant unit and troubleshooting without any specific interference in its operation[29].

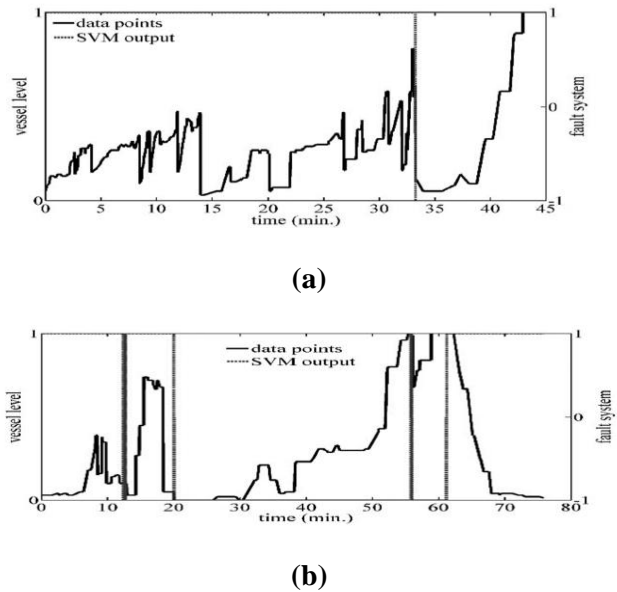

**(a)**

**(b)**

**Figure 10. (a) Start-up vessel level variations data for training and fault classification. (b) Fault detection system output and start-up vessel during test [29].**

## 4-4- An Example of solid mechanics applications

Andrejiova and colleagues published an article in 2018, named "Failure analysis of rubber composites under dynamic impact loading by logistic regression". Their main goal was to the monitoring of damage to rubber composite materials, particularly in the process of impact loading. Their test samples were extracted from conveyor belts of the same type (the same structure) and characterized four various phases of the conveyor belt life cycle (new, stored, worn, renovated). They applied the logistic regression method for analysis of the test samples damage significance. The purpose of the analysis was to find a model to describe the relationships between a categorical outcome variable and one or more categorical or continuous predictor variables. They found out that the highest odds of significant damage are in the case of stored conveyor belts. They also indicated, Renovated, worn, and new conveyor belts have better properties, in terms of the odds of significant damage. In the case of sharp-edged materials and a worn conveyor belt showed the best results[30].

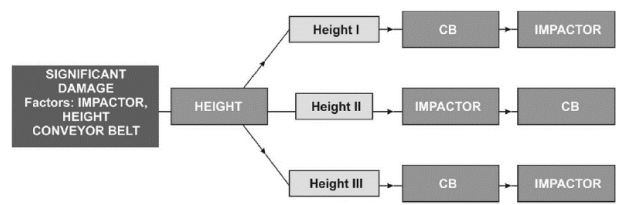

**Figure 11. The Decision tree scheme that was used [30].**

**Table 5. Estimates of the logistic model coefficients (α =0.05)[30].**

| Parameter | DF | Estimate | Standard error | Wald Chi-Square | p-Value |
|---|---|---|---|---|---|
| INTERCEPT ($\beta_0$) | 1 | − 13.40 | 3.03 | 19.56 | < 0.0001 |
| HEIGHT ($\beta_1$) | 1 | 5.59 | 1.27 | 19.37 | < 0.0001 |
| IMPACTOR ($\beta_2$) | 1 | 2.36 | 0.76 | 9.64 | 0.002 |
| STORED ($\beta_{21}$) | 1 | 2.04 | 1.02 | 4.00 | 0.046 |
| RENOVATED ($\beta_{22}$) | 1 | 0.16 | 0.92 | 0.03 | 0.862 |
| WORN ($\beta_{23}$) | 1 | − 0.55 | 0.91 | 0.36 | 0.549 |

The mentioned researches were just some examples of the applications of data science methods in mechanical engineering researches, which from their results, that is obvious using data science methods can proceed much precise and faster researches in modeling and simulation of mechanical systems and solving engineering problems studies.

## 5 . Conclusion

In this article, "Data Science" and its methods were reviewed briefly and simply, then discussed why mechanical engineers and researchers need to having knowledge in data science and using these methods in their works.

Machine learning is one of the crucial stages in data science and most of the researchers in mechanical engineering are using machine learning in their researches, so machine learning has an important role in their works, but it should be noted that this section alone may not perform well, and it is essential that researchers need to have a very good understanding of the rest of the steps of data science which been mentioned in the article for having a better results and decision making in their researches.

As mentioned, data science is constantly evolving and every day its applications are becoming more tangible and mechanical engineering is one of the areas that can be strongly linked to data science. Since, the most mechanical engineering issues have complexities, both numerically and conceptually, there may not always well be understood, and engineers and researchers may not be able to solve various problems easily and quickly, hence, they can get help from data science and solve complex problems and turn them to the simplest one.

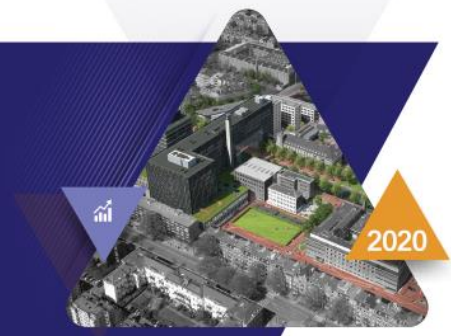